\documentclass[10pt,twocolumn,letterpaper]{article}

\usepackage[pagenumbers]{cvpr} 

\usepackage[dvipsnames]{xcolor}

\definecolor{cvprblue}{rgb}{0.21,0.49,0.74}
\usepackage[pagebackref,breaklinks,colorlinks,citecolor=cvprblue]{hyperref}

\def\paperID{-} 
\def\confName{CVPR}
\def\confYear{2025}

\usepackage{makecell}
\usepackage{graphicx}
\usepackage{amsmath, bm}
\usepackage{tabularx,booktabs}
\usepackage{adjustbox}
\usepackage{multirow}
\usepackage{graphicx,xcolor,colortbl} 
\definecolor{LightCyan}{rgb}{0.88,1,1}
\usepackage{array}
\newcolumntype{C}[1]{>{\centering\arraybackslash}p{#1}}
\newcolumntype{L}[1]{>{\arraybackslash}p{#1}}
\usepackage{indentfirst}
\usepackage{pifont}
\usepackage{xcolor}
\usepackage{hyperref}

\title{PatchRefiner V2: Fast and Lightweight Real-Domain \\ High-Resolution Metric Depth Estimation}

\author{%
Zhenyu Li, Wenqing Cui, Shariq Farooq Bhat, Peter Wonka \\
King Abdullah University of Science and Technology (KAUST) \\
\small\url{https://zhyever.github.io/patchrefinerv2/} \\
{\tt\small zhenyu.li.1@kaust.edu.sa}
}

\begin{document}
\maketitle
\begin{abstract}
While current high-resolution depth estimation methods achieve strong results, they often suffer from computational inefficiencies due to reliance on heavyweight models and multiple inference steps, increasing inference time. To address this, we introduce PatchRefiner V2 (PRV2), which replaces heavy refiner models with lightweight encoders. This reduces model size and inference time but introduces noisy features. To overcome this, we propose a Coarse-to-Fine (C2F) module with a Guided Denoising Unit for refining and denoising the refiner features and a Noisy Pretraining strategy to pretrain the refiner branch to fully exploit the potential of the lightweight refiner branch. Additionally, we introduce a Scale-and-Shift Invariant Gradient Matching (SSIGM) loss to enhance synthetic-to-real domain transfer. PRV2 outperforms state-of-the-art depth estimation methods on UnrealStereo4K in both accuracy and speed, using fewer parameters and faster inference. It also shows improved depth boundary delineation on real-world datasets like CityScape, ScanNet++, and KITTI, demonstrating its versatility across domains.
\end{abstract}    
\section{Introduction}
\label{sec:intro}

Accurate high-resolution depth estimation from a single image is critical for advancements in fields such as autonomous driving, augmented reality, and 3D reconstruction~\cite{eigen2014mde,zhang2023controlnet,bhat2023zoedepth,li2022binsformer}. Current state-of-the-art depth estimation models typically operate at relatively low resolutions (e.g. 0.3 megapixels). High memory requirements, especially at 4K resolution, pose a significant challenge for training depth estimation models that can natively support high-resolution inputs. Recent 4K depth estimation approaches like PatchRefiner~\cite{li2024patchrefiner} (PR) use a tile-based strategy where the high-resolution image is divided into patches. The patch-level depth predictions (fine, local outputs) are then fused with the depth prediction of a downsampled version of the input image (coarse, global output) to obtain a single, consistent, high-resolution output.

Despite its success, the PatchRefiner framework faces critical computational efficiency and scalability challenges for real-world applications. It employs the same architecture (a pre-trained base depth model) to extract both global as well as patch-level features. This amounts to at least 17 forward passes of the base model for a single high-resolution input. As the base model used~\cite{bhat2023zoedepth,yang2024depthanything,yang2024depthanythingv2} is often large, this results in two major issues: 1) \textbf{High inference time} of more than a second per image, and more importantly 2) \textbf{High memory requirement}, making the end-to-end training infeasible. Therefore, the PR framework has to adopt stage-wise training, where global and local branches are trained sequentially, leading to a long training time and suboptimal results.

\begin{figure}[t]
	\includegraphics[width=0.99\columnwidth]{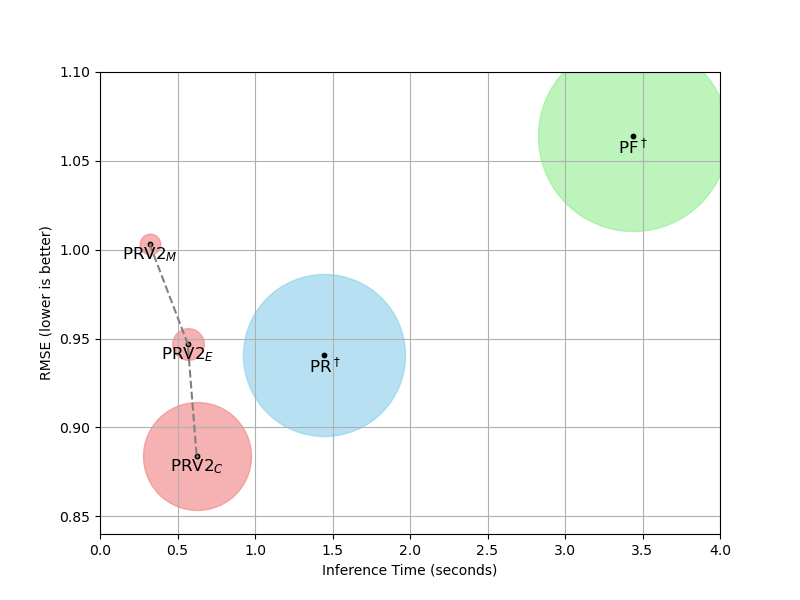}
	\hspace{-44mm}
        \scalebox{0.6}{
		\begin{tabular}[b]{l|ccr}
			& RMSE & \#params  & \multicolumn{1}{c}{T}\\
            \hline
			PF$^\dagger$ & 1.064 & 432.7M & 3.44s\\
            PR$^\dagger$ & 0.941 & 369.0M & 1.45s\\
            \hline
            PRV2$_{M}$ & 1.003 & \textbf{47.0M} & \textbf{0.32s} \\
            PRV2$_{E}$ & 0.948 & 72.1M & 0.57s \\
            PRV2$_{C}$ & \textbf{0.884} & 245.8M & 0.62s \\
            \multicolumn{3}{l}{~~$^\dagger$Aligned Version ~~\vspace{11mm} } \\
	\end{tabular}}
	\caption{\textbf{UnrealStereo4K results.} PatchRefiner V2 (PRV2) significantly outperforms previous high-resolution frameworks. In particular, PRV2$_{C}$ achieves new SOTA RMSE but being 2.3x faster than PR. PRV2$_{M}$ is 9.2x smaller and 10.7x faster than PF. PF and PR are short for PatchFusion~\cite{li2023patchfusion} and PatchRefiner~\cite{li2024patchrefiner}, respectively. We present the comparison of PR and PRV2 in Fig.~\ref{fig:arch_compare}.}
\label{fig:teaser}
\end{figure}

To alleviate these issues, we propose to substitute the large foundational models, such as ZoeDepth~\cite{bhat2023zoedepth} or DepthAnything~\cite{yang2024depthanything,yang2024depthanythingv2}, used in the refiner branch~\cite{li2024patchrefiner} with lightweight encoders like MobileNet~\cite{howard2019mobilenetv3,qin2024mobilenetv4} and EfficientNet~\cite{efficientnet}. This change significantly reduces the number of parameters and memory usage, decreases inference time, and enables end-to-end training without bells and whistles. 
However, this modification introduces a trade-off: the model capacity is reduced, and the refiner branch now lacks the depth-aligned feature representation otherwise provided by the previously used pre-trained depth estimation base models. While end-to-end training alleviates some of this limitation, the lack of depth-aligned feature representation remains a concern. Indeed, we observe that the features generated by these lightweight encoders tend to be `noisy' (see Fig.~\ref{fig:c2f_motivation}) even after ImageNet initialization~\cite{deng2009imagenet} and end-to-end training. This causes the original Fine-to-Coarse module (F2C) used in PR to struggle to inject rich high-resolution information for the final depth prediction. 


We propose two components to improve the feature representation in the refiner branch: 1) The \textbf{Coarse-to-fine module (C2F)} which incorporates novel Guided Denoising Units (GDUs) in a bottom-to-top manner~\cite{lin2017refinenet,xian2018monocular,ranftl2021dpt}. GDUs utilize coarse depth features as guidance to denoise and enhance the high-resolution refiner features. Together with the original Fine-to-Coarse module (F2C), this establishes a bidirectional fusion process: C2F initially denoises and refines high-resolution features using coarse features, followed by F2C’s enhancement of the predicted coarse depth map via residual prediction. 2) \textbf{Noisy Pre-training.}\footnote{We use the term `pre-training' loosely, as this process occurs prior to the final training phase.} Given that the C2F and F2C modules require initialization from scratch, we propose a simple pre-training strategy for the entire refiner branch — including the encoder, C2F, and F2C modules — to enhance feature representation and accelerate learning. During Noisy Pre-training, we replace the input coarse depth features for GDUs with random noise, essentially forcing the refiner branch to learn to extract depth-relevant features from the high-resolution input (Sec.~\ref{subsubsec:prv2:ap}).

\begin{figure}
    \centering
        \begin{tabular}{c}
             \includegraphics[width=0.75\linewidth]{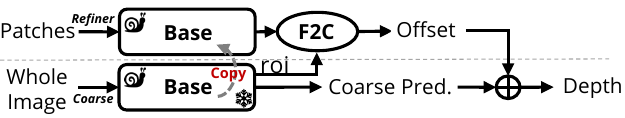}\vspace{-0.1cm}\\
             \footnotesize (a) PatchRefiner V1 Framework \vspace{0.3cm}\\
             \includegraphics[width=0.75\linewidth]{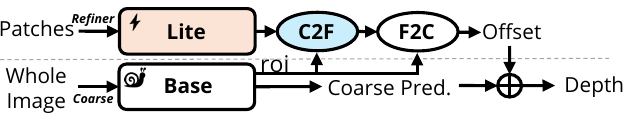}\vspace{-0.1cm}\\
             \footnotesize (b) Our PatchRefiner V2 Framework \vspace{-0.2cm}\\
        \end{tabular}
    \caption{A comparison of (a) PatchRefiner and (b) our proposed PatchRefiner V2. We adopt a lightweight encoder for the refiner branch, which alleviates the inference speed bottleneck, reduces the number of parameters for high-resolution estimation, and facilitates end-to-end training. A novel coarse-to-fine (C2F) module is proposed to denoise features from the lite model and further boost performance.}
    \label{fig:arch_compare}
\end{figure}

Finally, the PR framework~\cite{li2024patchrefiner} employs the Detail and Scale Disentangling (DSD) training strategy to adapt the high-resolution depth estimation framework to real-domain datasets, which enables learning `detail' from synthetic data and `scale' from the real domain. To isolate the scale from the synthetic data, the DSD strategy uses a ranking loss and Scale-and-Shift Invariant (SSI) loss~\cite{Ranftl2022midas}. 
Motivated by \cite{li2018megadepth}, we propose to replace the SSI loss with the Scale-and-Shift Invariant Gradient Matching (SSIGM) loss to learn high-frequency details from the synthetic data directly. 

Experiments demonstrate that our advanced framework, \textbf{PatchRefiner V2} (PRV2), performs effectively across various lightweight architectures. As summarized in Fig.~\ref{fig:teaser}, PRV2 significantly outperforms other high-resolution metric depth estimation frameworks on the UnrealStereo4K\cite{tosi2021smd} dataset in terms of both quantitative results and inference speed. Additionally, we evaluate the effectiveness of our adopted SSIGM loss across various frameworks (PR and PRV2) on diverse real-world datasets, including CityScape~\cite{cordts2016cityscapes} (outdoor, stereo), ScanNet++\cite{yeshwanth2023scannet++} (indoor, LiDAR and reconstruction), and KITTI\cite{schops2017eth3d} (outdoor, LiDAR). Our method reveals significant improvements in depth boundary delineation (e.g., +17.2\% boundary F1 on CityScape \textit{w.r.t} \cite{li2024patchrefiner}) while maintaining accurate scale estimation, showcasing its adaptability and effectiveness across different datasets and domains.


\section{Related Work}
\label{sec:related}

\subsection{High-Resolution Monocular Depth Estimation}

Monocular depth estimation (MDE) is a fundamental computer vision task and has recently seen impressive progress with advanced network design~\cite{eigen2014mde,bhat2021adabins,li2023depthformer,li2022binsformer,bhat2023zoedepth,yang2021transformer}, supervision~\cite{lee2020multiloss,liu2023va,xian2020rankloss,Ranftl2022midas,godard2019mde2}, formulation~\cite{fu2018dorn,diaz2019soft,bhat2021adabins,xian2020rankloss,li2022binsformer,bhat2022localbins}, training strategy~\cite{petrovai2022pseudomde,godard2019mde2,fan2023contrastive,Ranftl2022midas}, public datasets~\cite{silberman2012nyu,geiger2013kitti,dai2017scannet,cordts2016cityscapes,roberts2021hypersim}, \textit{etc}.
Recently, most SOTA frameworks~\cite{bhat2023zoedepth,yang2024depthanything,yang2024depthanythingv2,ke2024repurposing} build on the top of heavy backbones~\cite{bao2021beit,dosovitskiy2020vit,oquab2023dinov2,rombach2022sd}, leading to the limitation of low-resolution input. For example, Depth Anything V2~\cite{yang2024depthanythingv2} uses ViT-L~\cite{dosovitskiy2020vit,oquab2023dinov2} and can only infer 756$\times$994 (about 0.75 megapixels) images on an NVIDIA V100 32G GPU. While another line of research utilizing the generative model for MDE achieves fine-grained results, a similar dilemma exists. For instance, Marigold~\cite{ke2024repurposing} based on Stable Diffusion~\cite{rombach2022sd} runs with $\sim$0.33 megapixels as default.

\begin{figure*}
    \centering
    \setlength{\tabcolsep}{1pt}
        \begin{tabular}{cccc}
        
             \includegraphics[width=0.2\linewidth]{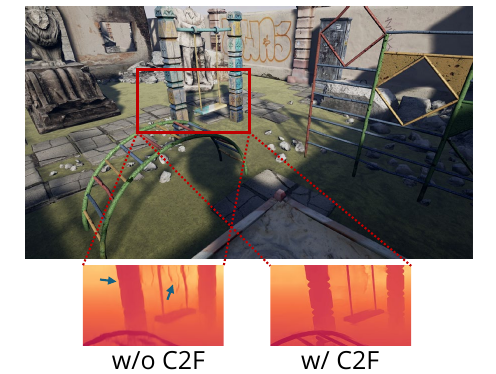}&
             \includegraphics[width=0.2\linewidth]{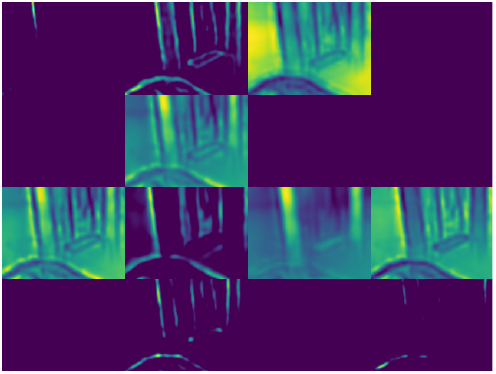}&
             \includegraphics[width=0.2\linewidth]{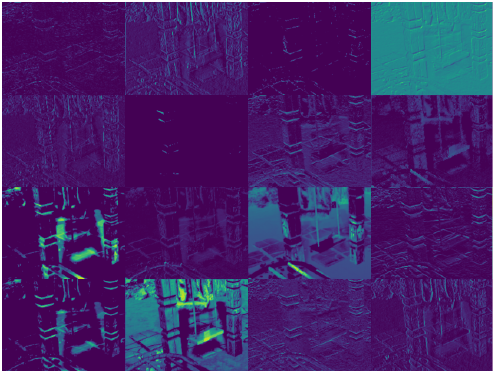}&
             \includegraphics[width=0.2\linewidth]{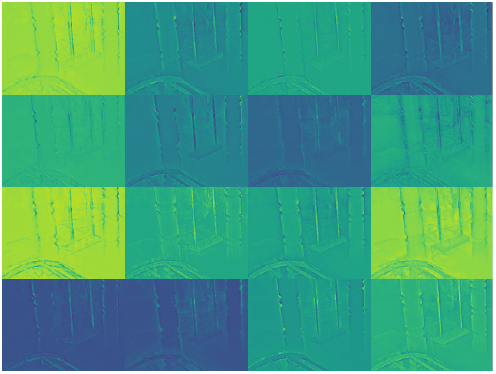}\\
             
             \footnotesize (a) Input, Prediction &
             \footnotesize (b) Coarse Feature &
             \footnotesize (c) Refiner Feature (w/o C2F) &
             \footnotesize (d) Refiner Feature (w/ C2F) \\
        \end{tabular}
    \caption{\textbf{Visualization of F2C input feature maps.} We showcase the first 16 channels of the F2C input features. (c) Without the C2F module (setting \ding{174} in Tab.~\ref{tab:arch_ablation}), the refiner features are `noisy' and hard to interpret. (d) The C2F module helps denoise the refiner features, leading to clear boundaries and better results.}
    \label{fig:c2f_motivation}
\end{figure*}

This contrasts with the advancements in modern imaging devices, which increasingly capture images at higher resolutions, reflecting the growing demand for high-resolution depth estimation~\cite{li2023patchfusion}. To relax the constraints, initial efforts utilize Guided Depth Super-Resolution (GDSR)~\cite{zhao2022dgsrdiscrete,metzger2023gdsrdiff,hui2016gdsrdepth,zhong2023guided} and Implicit Functions~\cite{mildenhall2021nerf,chen2021liif}. Recently, Tile-Based Methods have emerged as an alternative strategy~\cite{miangoleh2021boostingdepth,li2023patchfusion,li2024patchrefiner}, segmenting images into patches for individual estimation before reassembling them into a comprehensive depth map. Since all these methods adopt the dual branch architecture and utilize the same SOTA depth model in both branches, the frameworks are heavy and slow at inference time. By contrast, we aim to achieve fast high-resolution metric depth estimation using the tile-based method with fewer additional parameters.

\subsection{Synthetic-to-Real Transfer for MDE}

The challenge of collecting high-quality real-domain data for high-resolution depth training has prompted the use of synthetic datasets~\cite{rajpal2023high,li2024patchrefiner}. Different from traditional methods that utilize the synthetic data within unsupervised domain adaptation frameworks~\cite{chen2019crdoco,kundu2018adadepth,lopez2023desc,koutilya2020sharingan,zhao2019geometry,zheng2018t2net}, PatchRefiner proposes a more practical setting in which labeled data from both synthetic and real domains is leveraged to improve real-world, high-resolution depth estimation~\cite{li2024patchrefiner}. Motivated by the success of semi-supervised learning~\cite{van2020survey,yang2024depthanything,kirillov2023sam}, they adopt the pseudo-labeling strategy~\cite{pseudo2013simple,saito2017asymmetric,chen2019progressive,pastore2021closer,shin2022mm} with the Detail and Scale Disentangling (DSD) loss to transfer the fine-grained knowledge learned from the synthetic data to the real domain. However, the adopted scale-and-shift invariant (SSI) loss~\cite{Ranftl2022midas,yang2024depthanything} for detail supervision is indirect and can be ambiguous.
This work introduces scale-and-shift invariant \textit{gradient matching}~\cite{li2018megadepth} to achieve more effective synthetic-to-real transfer.

\begin{figure*}
    \centering
    \includegraphics[width=0.75\linewidth]{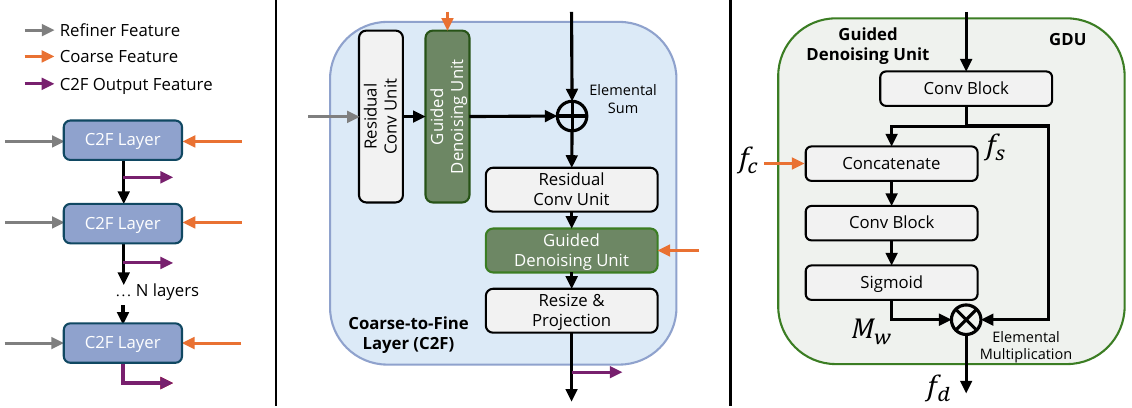}
    \caption{\textit{Left}: Coarse-to-Fine (C2F) module overview. It processes refiner features in a bottom-to-top manner with $N$ successive C2F layers. Each layer is guided by coarse features with corresponding resolution and outputs denoised features for the Fine-to-Coarse (F2C) module. \textit{Center}: C2F layers combine multi-level features with Residual Convolutional Units~\cite{lin2017refinenet,Ranftl2022midas} and denoises the features using Guided Denoising Units (GDU). \textit{Right}: Guidance information from the coarse branch is introduced through a concatenation followed by a convolutional block and then converted to a weight map ranging from 0 to 1 through the sigmoid operator. We then adopt an elementwise multiplication to denoise the shortcut feature.}
    \label{fig:overall_arch}
\end{figure*}

\section{Method}
\label{sec:method}


\subsection{Revisiting PatchRefiner V1}
\label{subsec:prv1}

We first revisit the PatchRefiner framework~\cite{li2024patchrefiner} (named as PR~V1). The PR~V1 framework adopts a tile-based approach to address the high memory and computational demands of high-resolution depth estimation~\cite{li2023patchfusion, miangoleh2021boostingdepth}. It utilizes a two-step process: \textbf{(i)} Coarse Depth Estimation and \textbf{(ii)} Fine-Grained Depth Refinement, as shown in Fig.~\ref{fig:arch_compare}.

\textbf{(i) Coarse Depth Estimation:} This step involves a coarse depth estimation network, $\mathcal{N}_{c}$, which processes downsampled inputs to generate a global depth map, $\mathbf{D}_{c}$. This map captures the overall scene structure and provides a baseline for further refinement. Notably, $\mathcal{N}_{c}$ can be any depth estimation model and is kept fixed after this stage.

\textbf{(ii) Fine-Grained Depth Refinement:} PR introduces a unified refinement network, $\mathcal{N}_{r}$, in place of separate fine depth networks and fusion mechanisms~\cite{li2023patchfusion, poucin2021boosting}. This network refines the coarse depth map by recovering details and enhancing depth precision at a patch level.

The refinement process begins with the cropped input image $I$, processed by a base depth model $\mathcal{N}_{d}$, which shares the same architecture as $\mathcal{N}_{c}$. Multi-scale features from both $\mathcal{N}_{d}$ and $\mathcal{N}_{c}$ are collected as $\mathcal{F}_d=\{f_d^i\}_{i=1}^{L}$ and $\mathcal{\Tilde{F}}_c=\{\Tilde{f}_c^i\}_{i=1}^{L}$. Following~\cite{li2023patchfusion}, the $\texttt{roi}$~\cite{he2017mask} operation extracts features from the cropped area as $\Tilde{f}_c^i=\texttt{roi}(f_c^i)$.

These features are then aggregated by a lightweight decoder through concatenation and convolutional blocks, referred to as the Fine-to-Coarse (F2C) module in this paper, which injects fine-grained information into the coarse refinement process. The F2C module constructs the residual depth map $\mathbf{D}_{r}$ at the input resolution, and the final patch-wise depth map is computed as $\mathbf{D} = \texttt{roi}(\mathbf{D}_{c}) + \mathbf{D}_{r}$.

As the second contribution, PR introduces a teacher-student framework to transfer the fine-grained knowledge learned from the synthetic data to the real domain. The Detail and Scale Disentangling (DSD) loss is designed to help the model balance detail enhancement with scale accuracy by integrating both the scale-consistent ground truth supervision and the detail-focused pseudo labels. Both ranking loss~\cite{xian2020rankloss} and the scale-and-shift invariant loss~\cite{Ranftl2022midas} can be adopted for pseudo-label supervision.

\paragraph{Limitations of PR~V1.}
Similar to other tile-based methods~\cite{poucin2021boosting,li2023patchfusion}, the PR framework encounters significant challenges with the computational efficiency and scalability for real-world applications due to the shared usage of the base depth model (e.g., ZoeDepth~\cite{bhat2023zoedepth}, Depth Anything~\cite{yang2024depthanything,yang2024depthanythingv2}) across both the coarse and refiner branches. For a given input image, while the coarse branch processes the downsampled image once to gather global information, the refiner branch requires multiple inferences (at least 16 in PR's default mode) for the patches. Since both branches share the same architecture, the refiner branch becomes the primary efficiency bottleneck. Our goal is to alleviate this bottleneck as much as possible.

Moreover, a heavy framework makes end-to-end training infeasible due to GPU memory limitations. The PR framework has to adopt two stages for training the framework, where global and local branches are trained sequentially. This results in a long training time and suboptimal performance. While the authors claim that multiple-stage training could potentially lead to stage-wise local optima~\cite{li2024patchrefiner}, our goal is to pursue end-to-end training.

\subsection{PatchRefiner V2 Framework}

\subsubsection{Lite Framework for Faster Inference and End-to-End Training}
\label{subsubsec:prv2:speed}
We propose a simple solution to address PR~V1's limitations: a lightweight architecture for the refiner branch. Given that the coarse branch already provides a reliable base depth estimation $\mathbf{D}_{c}$, using the same heavy model for the refiner might be unnecessary. This substitution significantly increases inference speed, reduces the model size, and enables end-to-end training. However, it also results in a noticeable decline in refinement quality compared to previous methods~\cite{li2023patchfusion,li2024patchrefiner}. We attribute this decline to the lack of depth-aligned feature representation in the refiner branch, as shown in Fig.~\ref{fig:c2f_motivation}. 

To compensate for the loss in model capacity and depth-pretraining by the proposed substitution, we introduce a better architecture design, a Coarse-to-Fine (C2F) Module, and a fast and simple pre-training strategy, Noisy Pretraining (NP).

\begin{figure*}[t]
\setlength\tabcolsep{0.1pt}
\centering
\small
    \begin{tabular}{@{}*{4}{C{3.7cm}}@{}}
    \includegraphics[width=1\linewidth]{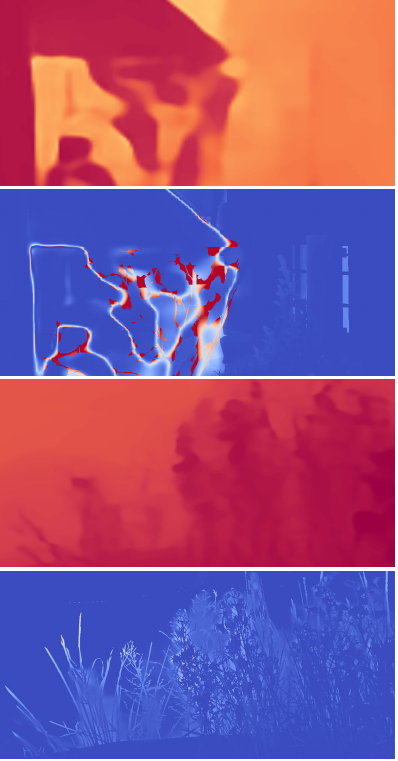} &
    \includegraphics[width=1\linewidth]{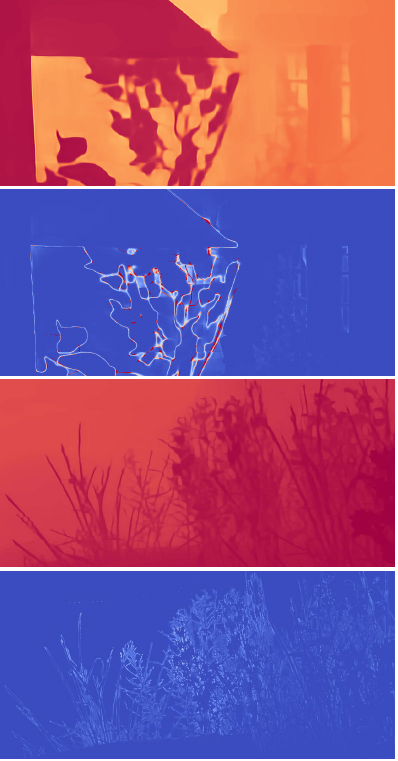} &
    \includegraphics[width=1\linewidth]{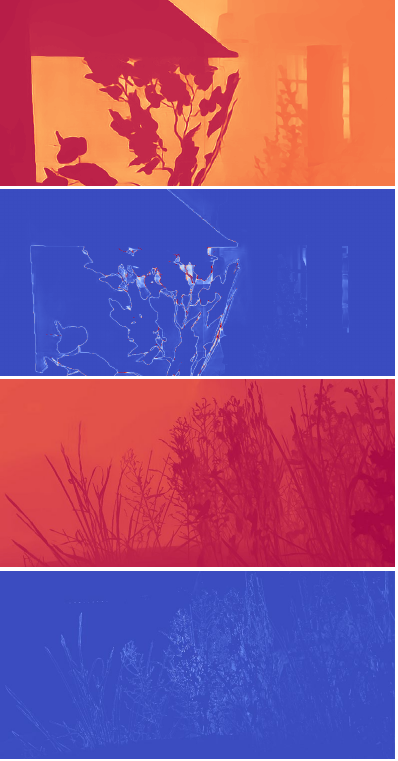} &
    \includegraphics[width=1\linewidth]{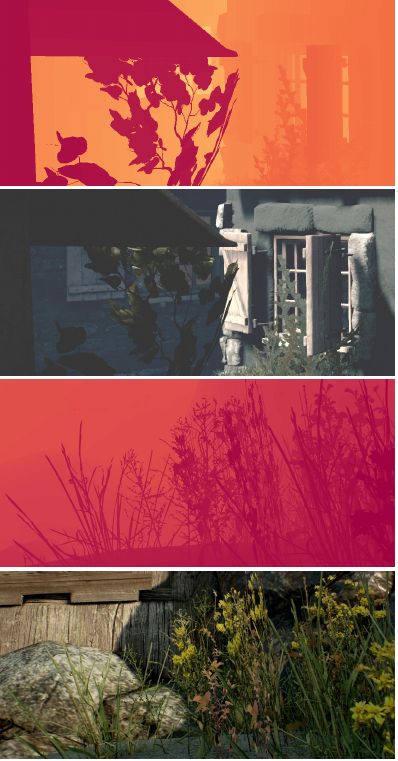} \\
    
    ZoeDepth~\cite{bhat2023zoedepth} & PatchRefiner~\cite{li2023patchfusion} & PRV2$_{\textsc{C}}$ & GT, Input \\
    \end{tabular}
    \caption{\textbf{Qualitative Comparison on UnrealStereo4K.} We show the depth prediction and corresponding error map, respectively. The qualitative comparisons showcased here indicate our PRV2$_{\textsc{C}}$ outperforms counterparts~\cite{bhat2023zoedepth,li2024patchrefiner} with sharper edges and lower error around boundaries while achieving faster inference. We show individual patches in all images to emphasize details near depth boundaries.}
    \label{fig:u4k}
\end{figure*}

\subsubsection{Coarse-to-Fine Module}
\label{subsubsec:prv2:c2f}

Since the refiner branch no longer includes a pretrained depth model, we propose utilizing information from the global coarse branch to guide the selection of relevant details from the fine, patch-level features.

The proposed Coarse-to-Fine (C2F) module shown in Fig.~\ref{fig:overall_arch} processes the multi-scale features extracted from the lightweight encoder through $N$ successive C2F layers in a bottom-to-up manner~\cite{ronneberger2015unet,lin2017refinenet}, mirroring the design of the Fine-to-Coarse (F2C) module~\cite{li2024patchrefiner}. Each C2F layer is designed to progressively enhance and denoise the refiner features with the help of coarse feature representations.

Each layer in the C2F module consists of two components: our proposed Guided Denoising Unit (GDU) and the Residual Convolutional Unit~\cite{lin2017refinenet,Ranftl2022midas}. The GDU introduces coarse feature maps $f_c$ at each stage to refine and denoise the refiner features. Specifically, the coarse features serve as guidance, which are incorporated via the concatenation operation ($\texttt{Cat}$) followed by the convolutional block ($\texttt{CB}$). The output of these blocks is passed through a sigmoid activation function ($\sigma$) to obtain a weight map $M_{w}$, which ranges from 0 to 1. This weight map is then applied to the shortcut features $f_{s}$ through elemental multiplication $\otimes$, effectively denoising the shortcut features. This process can be formulated as
\begin{equation}
    M_{w} = \sigma(\texttt{CB}(\texttt{Cat}(f_c,f_s))),
\end{equation}
\begin{equation}
    f_d = M_{w} \otimes f_s,
\end{equation}

\noindent where the $f_d$ indicates the denoised feature. Associated with the Residual Convolutional Unit, it allows the model to filter out irrelevant noise and enhance the quality of the refined features iteratively across the network layers. After that, we utilize the F2C module to inject the denoised fine-grained information for coarse features, leading to a more effective and better refinement process.

\subsubsection{Noisy Pretraining}
\label{subsubsec:prv2:ap}

In PR~V1, the framework's efficacy largely depends on the comprehensive pretraining of the base models in both the coarse and refiner branches~\cite{li2024patchrefiner}. During the subsequent high-resolution training stage, only the Fine-to-Coarse (F2C) module is trained from scratch, representing a minor portion of the overall refiner branch (24.0M \textit{vs.} 369.0M parameters). In other words, a significant portion ($\sim$94\%) of the refiner branch is pretrained for depth estimation.

By our substitution, this pretraining is also lost. While the lightweight encoder used can be pretrained on a large-scale dataset with complex strategies, it now constitutes only a small part of the refiner branch (1.3M \textit{vs.} 47.0M parameters for PRV2$_{\text{M}}$) and lacks the depth-aligned feature representation. In other words, even if we pre-train the encoder, a significant portion ($\sim$98\%) of the refiner branch must still be trained from scratch. 
%

To address this issue, we propose a novel approach called Noisy Pretraining (NP). Prior to the high-resolution training, we pretrain the lightweight encoder along with the C2F and F2C modules. However, a critical aspect of our framework is that both the C2F and F2C modules rely on features from the base model in the coarse branch. These features, however, are challenging to omit during the pretraining process. We propose a straightforward yet effective solution: we randomly generate the coarse features using a normal distribution $N(0,1)$ as inputs. This forces the refiner branch to learn depth-relevant features without guidance from the coarse branch.

Unlike other strategies~\cite{liu2023zero, ozguroglu2024pix2gestalt, brooks2023instructpix2pix}, which often require careful selection and modification of convolutional layers and their corresponding parameters, our NP method avoids altering the framework’s architecture. As a result, the pretraining and subsequent training stages proceed seamlessly, preserving the integrity of the overall model structure while ensuring that all components of the refiner branch are well-prepared for high-resolution training.

\begin{table*}[t!]
    \centering
    \scalebox{0.7}{
    \begin{tabular}{L{3.3cm}|*{5}{C{1.5cm}}|*{2}{C{1.5cm}}|C{2.2cm}}
        \toprule
        Method & \boldsymbol{$\delta_1 (\%)$}$\uparrow$ & \textbf{REL}$\downarrow$ & \textbf{RMS}$\downarrow$ & \textbf{SiLog}$\downarrow$ & \textbf{SEE}$\downarrow$ & \textbf{\#param}$\downarrow$ & T$\downarrow$ & Reference \\
        \midrule
        iDisc~\cite{piccinelli2023idisc}          & 96.940 & 0.053 & 1.404 & 8.502 & 1.070 & \multirow{4}{*}{\textbf{-}} & \multirow{4}{*}{\textbf{-}} & ICCV 2023 \\
        SMD-Net~\cite{tosi2021smd} & 97.774 & 0.044 & 1.282 & 7.389 &  0.883 & & & CVPR 2021  \\
        Graph-GDSR~\cite{de2022gdsr} & 97.932 & 0.044 & 1.264 & 7.469 & 0.872 & & & CVPR 2022 \\
        BoostingDepth~\cite{miangoleh2021boostingdepth} & 98.104 & 0.044 & 1.123 & 6.662 & 0.939 & & & CVPR 2021 \\
        \midrule  
        ZoeDepth~\cite{bhat2023zoedepth}  & 97.717 & 0.046 & 1.289 & 7.448 & 0.914 & - & - & - \\ 
        \midrule
        \midrule  
        \textcolor{gray}{ZoeDepth+PF}~\cite{li2023patchfusion} & \textcolor{gray}{98.419} & \textcolor{gray}{0.040} & \textcolor{gray}{1.088} & \textcolor{gray}{6.212} & \textcolor{gray}{0.838} & \multirow{2}{*}{432.7M} & \multirow{2}{*}{3.44s} & \multirow{2}{*}{CVPR 2024} \\
        ZoeDepth+PF$^\dagger$~\cite{li2023patchfusion} & 98.369 & 0.039 & 1.064 & 6.342 & 0.855 & & \\
        \midrule
        \textcolor{gray}{ZoeDepth+PR}~\cite{li2024patchrefiner} & \textcolor{gray}{98.821} & \textcolor{gray}{0.033} & \textcolor{gray}{0.892} & \textcolor{gray}{5.417} & \textcolor{gray}{\textbf{0.750}} & \multirow{2}{*}{369.0M} & \multirow{2}{*}{1.45s} & \multirow{2}{*}{ECCV 2024} \\ 
        ZoeDepth+PR$^\dagger$~\cite{li2024patchrefiner} & 98.680 & 0.034 & 0.941 & 5.614 & 0.771 & & \\
        \midrule
        ZoeDepth+\textbf{PRV2$_{\textsc{M}}$} & 98.610 & 0.034 & 1.003 & 5.760 & 0.832 & \textbf{47.0M} & \textbf{0.32s} & \multirow{3}{*}{\textbf{Ours}} \\ 
        ZoeDepth+\textbf{PRV2$_{\textsc{E}}$} & 98.728 & 0.034 & 0.948 & 5.579 & 0.816 & 72.1M & 0.57s & \\ 
        ZoeDepth+\textbf{PRV2$_{\textsc{C}}$} & \textbf{98.863} & \textbf{0.032} & \textbf{0.884} & \textbf{5.281} & 0.787 & 245.8M & 0.62s \\ 
        \bottomrule
    \end{tabular}
    }
    \caption{\textbf{Quantitative comparison on UnrealStereo4K.} Best results are marked \textbf{bold}. PF, PR and PRV2 are short for PatchFusion~\cite{li2023patchfusion}, PatchRefiner~\cite{li2024patchrefiner} and PatchRefiner V2, respectively. We report the $P=16$ mode for these high-resolution depth estimation frameworks~\cite{li2023patchfusion}. \textcolor{gray}{\textbf{Gray lines}} present numbers from the original paper with vanilla pretraining settings. $^\dagger$: indicates the pretraining aligned version, where we remove the \textbf{non-public} Midas pretraining stage~\cite{Ranftl2022midas} adopted for the \textit{fine or refiner branch} in PR and PF to make fair comparisons with our PRV2. The coarse branch is \textbf{NOT} modified. \#param. and T denote the number of additional parameters adopted for high resolution estimation and the inference time of the \textit{fine or refiner branch} for one input image. Best results are in \textbf{bold}.}
    \label{tab:arch}
\end{table*}

\subsection{Scale-and-Shift Invariant Gradient Matching}
\label{subsec:ssigm}

In the synthetic-to-real transfer stage, PR adopts the scale-and-shift invariant (SSI) loss $\mathcal{L}_{ssi}$ as the pseudo-label supervision as a part of the Detail and Scale Disentangling (DSD) loss $\mathcal{L}_{DSD}$.

Given a predicted depth $d_i$ and the corresponding pseudo label $\hat{d}_i$, the SSI loss first aligns them with the least-squares estimation (LSE) as:
\begin{gather}
    (s,t) = \mathrm{argmin}_{s,t} \sum\limits^M_{i=1} (sd_i + t - \hat{d_i})^2,\\
    d^* = sd + t,~ \hat{d}^*=\hat{d} 
\end{gather}
\noindent where the scale $s$ and shift $t$ factors are effectively determined with the closed form~\cite{Ranftl2022midas}. $d_i^*$ and $\hat{d}_i^*$ are scaled and shifted versions of the predicted depth and pseudo label, respectively. Then, a pixel-wise mean absolute error loss is calculated as
\begin{equation}
    \mathcal{L}_{ssi} = \frac{1}{M} \sum\limits_{i=1}^{M}\rho(d_i^*-\hat{d}_i^*),
\end{equation}
\noindent where $\rho$ is the mean absolute error (MAE) loss, and $M$ is the number of pixels. In this paper, we replace the MAE loss with the gradient matching loss~\cite{li2018megadepth} that applies a constraint on the gradient variation. It facilitates the model to learn the high-frequency details from the synthetic data directly~\cite{li2018megadepth}. We name this combination scale-and-shift invariant gradient matching (SSIGM) loss. It can be formulated as 
\begin{equation}
  \mathcal{L}_{ssigm} = \frac{1}{M} \sum_{i=1}^{M} \big(|\nabla_x R_i|+|\nabla_y R_i|\big),
  \label{eq:ssigm}
\end{equation}
where $R_i = \hat d_i- \hat{d}_i^*$. Similar to PR~\cite{li2024patchrefiner}, we utilize $\lambda$ to control the strength of pseudo-label supervision.

\section{Experiments}
\label{sec:exps}
\begin{figure*}[t]
\setlength\tabcolsep{0.1pt}
\centering
\small
    \begin{tabular}{@{}*{5}{C{2.7cm}}@{}}
    \includegraphics[width=1\linewidth]{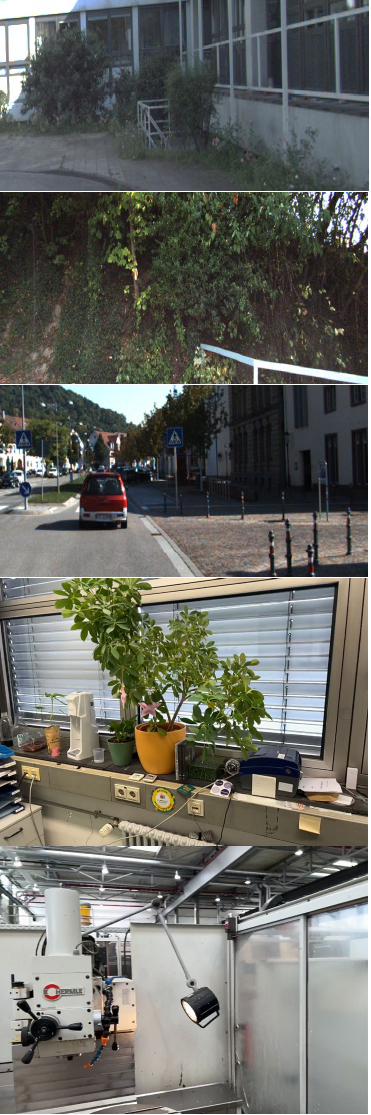} &
    \includegraphics[width=1\linewidth]{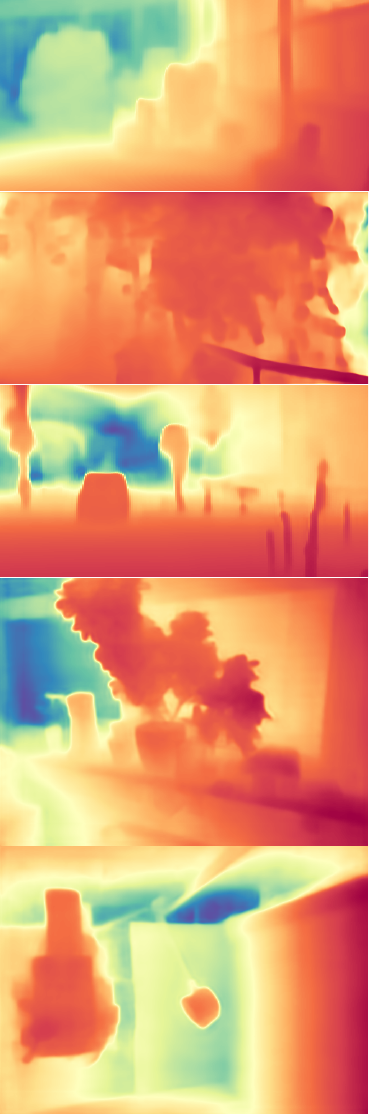} &
    \includegraphics[width=1\linewidth]{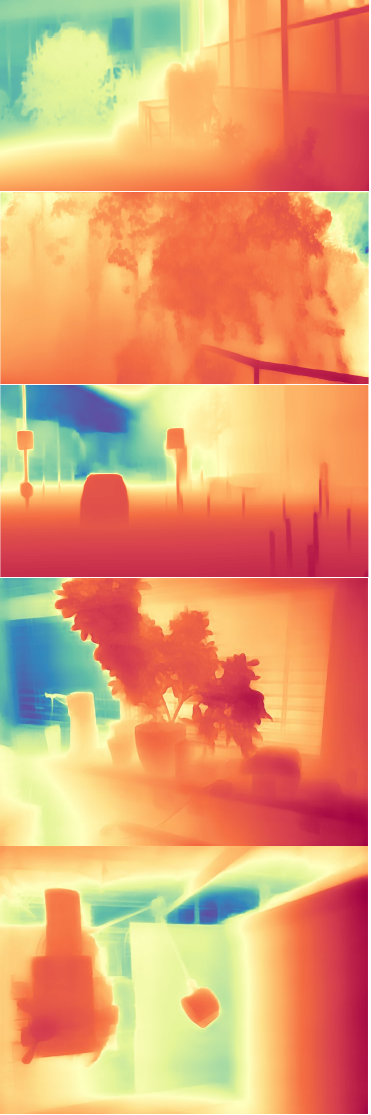} &
    \includegraphics[width=1\linewidth]{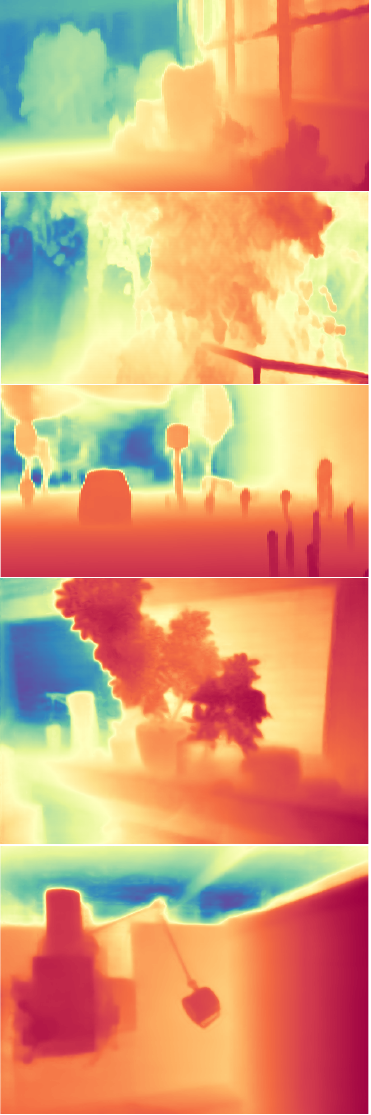} &
    \includegraphics[width=1\linewidth]{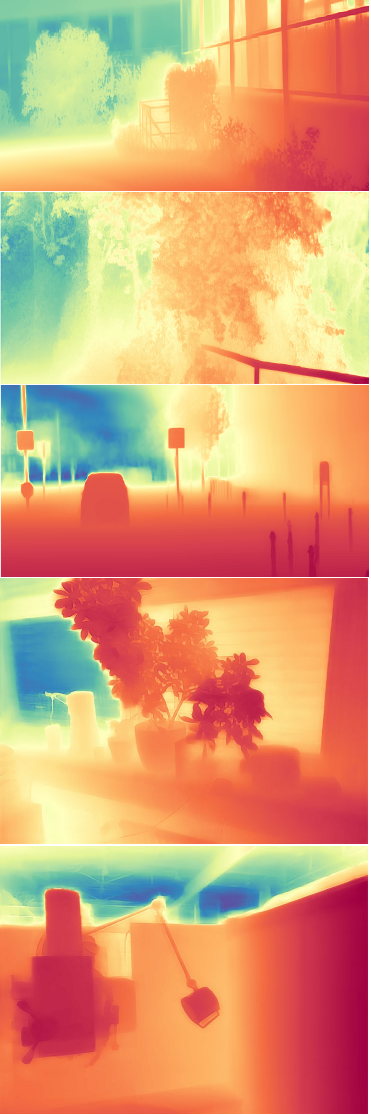} \\
    
    Input & ZoeDepth~\cite{bhat2023zoedepth} & Zoe+PRV2$_E$ & Depth Anything V2 & DAV2+PRV2$_E$ \\
    \end{tabular}
    \caption{\textbf{Qualitative Comparison on KITTI and ScanNet++.} Our PVR2 can consistently boost high-resolution depth estimation with various base models (ZoeDepth, DAV2) on various real-domain datasets. Zoom in to better perceive details near boundaries.}
    \label{fig:real}
\end{figure*}

\begin{table}[t!]
    \centering
    \scalebox{0.85}{
    \begin{tabular}{L{3.2cm}|*{3}{C{1.4cm}}}
        \toprule
        Method & \boldsymbol{$\delta_1 (\%)$}$\uparrow$ & \textbf{RMS}$\downarrow$ & \textbf{SEE}$\downarrow$ \\
        \midrule
        Depth Anything V2   & \multirow{2}{*}{98.072} & \multirow{2}{*}{1.144} & \multirow{2}{*}{0.908} \\ 
        (DAV2)~\cite{yang2024depthanythingv2} \\
        \midrule
        DAV2+\textbf{PRV2$_{\textsc{M}}$} & 98.771 & 0.919 & 0.813 \\ 
        DAV2+\textbf{PRV2$_{\textsc{E}}$} & 98.874 & 0.869 & 0.784 \\ 
        DAV2+\textbf{PRV2$_{\textsc{C}}$} & \textbf{98.929} & \textbf{0.859} & \textbf{0.779}  \\ 
        \bottomrule
    \end{tabular}
    }
    \caption{\textbf{Framework performance with Depth Anything V2 as the base model on UnrealStereo4K.} Our advanced framework can consistently boost the SoTA depth estimator for high-resolution metric depth estimation.}
    \label{tab:dav2}
\end{table}

\begin{table*}[t!]
    \centering
    \scalebox{0.7}{
    \begin{tabular}{L{4.2cm}|*{2}{C{0.5cm}}|*{1}{C{1.3cm}}|*{3}{C{1.6cm}}|*{3}{C{1.6cm}}}
        \toprule
        \multirow{2}{*}{Method} & \multicolumn{2}{c|}{Data} & \multirow{2}{*}{pl gen} & \multicolumn{3}{c|}{Scale} & \multicolumn{3}{c}{Boundary} \\
        \cmidrule{2-3}
        \cmidrule{5-10}
        & $\mathcal{S}$ & $\mathcal{R}$ & & \boldsymbol{$\delta_1 (\%)$}$\uparrow$ &  \textbf{REL}$\downarrow$ & \textbf{RMS}$\downarrow$  & \textbf{EdgeAcc}$\downarrow$ & \textbf{EdgeComp}$\downarrow$ & \textbf{F1}$\uparrow$  \\
        \midrule
        ZoeDepth~\cite{bhat2023zoedepth} &  & \checkmark & \multirow{2}{*}{-} & 92.214 & 0.081 & 9.097 & 3.62 & 42.18 & 19.15  \\ 
        ZoeDepth~\cite{bhat2023zoedepth} + FT & \checkmark & \checkmark & & 92.249 & 0.081 & 9.092 & 3.61 & 38.69 & 20.02 \\
        \midrule
        PR (zero-shot)~\cite{li2024patchrefiner} & \checkmark &   & - & 3.630 & 0.436 & 24.506  & 3.25 & \textbf{9.16} & 31.10 \\
        PR~\cite{li2024patchrefiner} &   &  \checkmark & - & 93.295 & 0.076 & 8.324 & 3.39 & 22.14 & 26.99\\
        PR + Ranking~\cite{li2024patchrefiner} & \checkmark & \checkmark & online & 93.308 & 0.076 & 8.313 & 3.28 & 17.65 & 29.53 \\
        PR + SSI~\cite{li2024patchrefiner}& \checkmark & \checkmark & online & 93.300 & 0.075 & 8.313 & 3.25 & 17.11 & 30.02 \\ 
        PR + \textbf{SSIGM} (Ours) & \checkmark & \checkmark & online & 93.277 & 0.076 & 8.315 & \textbf{3.00} & 15.47 & \textbf{35.19} \\ 
        PR + \textbf{SSIGM} (Ours) & \checkmark & \checkmark & offline & 93.320 & 0.076 & 8.297 & \underline{3.03} & \underline{14.70} & \underline{34.89}\\ 
        \midrule
        PRV2$_{\textsc{E}}$ &   &  \checkmark & - & 93.023 & 0.076 & 8.513 & 3.29 & 22.60 & 27.98 \\
        PRV2$_{\textsc{E}}$ + Ranking~\cite{li2024patchrefiner} & \checkmark & \checkmark & online & 92.905 & 0.077 & 8.533 & 3.30 & 21.72 & 28.27 \\
        PRV2$_{\textsc{E}}$ + SSI~\cite{li2024patchrefiner}& \checkmark & \checkmark & online & 92.901 & 0.076 &  8.533 & 3.24 & 21.12 & 29.22 \\ 
        PRV2$_{\textsc{E}}$ + \textbf{SSIGM} (Ours) & \checkmark & \checkmark & online & 92.883 & 0.077 & 8.534 & \textbf{3.04} & \underline{17.78} & \textbf{35.32} \\ 
        PRV2$_{\textsc{E}}$ + \textbf{SSIGM} (Ours) & \checkmark & \checkmark & offline & 92.889 & 0.077 & 8.530 & \underline{3.10} & \textbf{16.97} & \underline{34.85} \\ 
        \bottomrule
    \end{tabular}
    }
    \caption{\textbf{Quantitative comparison on CityScapes.} FT is short for fine-tuning. We mainly compare our proposed SSIGM loss with the ranking~\cite{xian2020rankloss} and scale-and-shift loss~\cite{Ranftl2022midas,yang2024depthanything} adopted in \cite{li2024patchrefiner}. Reported results are from the official implementation\protect\footnotemark. Best boundary results are in \textbf{bold}, second best are \underline{underlined} for each different framework.}
    \label{tab:s2r}

\end{table*}

\footnotetext{\url{https://github.com/zhyever/PatchRefiner/blob/main/docs/user_training.md}}

\subsection{Datasets and Metrics}
\noindent \textbf{Datasets:} We evaluate the effectiveness of our proposed framework on the UnrealStereo4K dataset~\cite{tosi2021smd} (Synthetic), which offers synthetic stereo images at a 4K resolution (2160$\times$3840), each paired with accurate, boundary-complete pixel-wise ground truth. Adhering to the dataset splits in~\cite{tosi2021smd,li2023patchfusion, li2024patchrefiner}, we employ a default patch size of 540$\times$960 for compatibility with~\cite{li2023patchfusion,li2024patchrefiner}. In terms of the synthetic-to-real transfer part, we use the Cityscapes~\cite{cordts2016cityscapes} (Real, Stereo), ScanNet++ (Real, LiDAR, Reconstruction), and KITTI (Real, LiDAR) datasets. Since the Cityscapes dataset offers a comprehensive suite of urban scene images, segmentation masks, and disparity maps at a relatively high resolution, we conduct comparison experiments on the Cityscapes dataset and provide qualitative comparison results on ScanNet++ and KITTI datasets. More details about datasets are provided in \textit{supplementary materials}. 

\noindent \textbf{Metrics:} Following \cite{li2023patchfusion,li2024patchrefiner}, we adopt standard depth evaluation metrics from~\cite{eigen2014mde,piccinelli2023idisc,bhat2023zoedepth} and the Soft Edge Error (SEE) from~\cite{tosi2021smd,chen2019over,li2023patchfusion} for \textit{scale} evaluation. As for the \textit{boundary} evaluation on the real-domain datasets (Cityscapes), we adopt the standard protocol introduced in \cite{li2024patchrefiner} and utilize the F1 score, Edge Accuracy, and Edge Completeness metrics to evaluate the boundary quality.

\subsection{Implementation Details}

\noindent \textbf{PRV2 on Synthetic Dataset:} For training on the synthetic dataset, we employ the scale-invariant log loss $\mathcal{L}_{silog}$, as introduced in~\cite{eigen2014mde,bhat2021adabins,lee2019big}. We initialize the coarse network $\mathcal{N}_c$ with pretrained weights from the NYU-v2 dataset~\cite{silberman2012nyu}, adhering to the approach in~\cite{li2023patchfusion,li2024patchrefiner} for a fair comparison. As for the refiner brach, we employ the MobileNet-Small~\cite{qin2024mobilenetv4}, EfficientNet-B5~\cite{efficientnet}, and Convnext-Large for PRV2$_M$, PRV2$_E$, and PRV2$_C$, respectively. We perform the noisy pretraining for the refiner branch for 96 epochs. The $\mathcal{N}_c$ is independently trained for 24 epochs and fine-tuned with the refiner branch in a fully end-to-end manner for another 48 epochs on the synthetic dataset. During inference, we implement Consistency-Aware Inference, as described in~\cite{li2023patchfusion}, to optimize performance.

\noindent \textbf{Learning on Real-Domain Dataset:} Following \cite{li2024patchrefiner}, we first train the entire framework on the target real-domain dataset with the same setting as the synthetic dataset. After that, we fine-tune the model with the Detail and Scale Disentangling loss for two epochs to refine depth estimations. To achieve a fair comparison with PR, we use the ZoeDepth+PR and DepthAnything V2+PR as the teacher model to generate pseudo labels for our ZoeDepth+PRV2 and DepthAnything V2+PRV2 models, respectively.

\subsection{Main Results}

\noindent \textbf{Synthetic Dataset:} As shown in Tab.~\ref{tab:arch}, our most lightweight model, PRV2$_{M}$, not only improves RMSE by 22.2\% compared to the base depth model but is also 9.2x smaller and 10.7x faster than PatchFusion (PF) in terms of parameter count and inference speed, respectively. Our middleweight model, PRV2$_{E}$, achieves comparable RMSE to the previous SoTA PR while being 2.5x faster and 5.1x smaller, offering an excellent balance between performance and efficiency. With ConvNext as the backbone, PRV2$_{C}$ sets a new SoTA with an RMSE of 0.884, while being 2.3x faster than PR. Qualitative results in Fig.~\ref{fig:u4k} demonstrate PRV2$_{C}$'s superior boundary delineation. Furthermore, when using Depth Anything V2 as the base model, as shown in Tab.~\ref{tab:dav2}, our framework consistently boosts performance across different lightweight refiner models, highlighting its versatility.

\noindent \textbf{Real-Domain Dataset:} Tab.~\ref{tab:s2r} and Fig.~\ref{fig:real} illustrate the performance gains achieved with our proposed SSIGM loss. While maintaining a comparable scale RMSE to the Ranking and SSI losses used in \cite{li2024patchrefiner}, SSIGM significantly improves boundary F1 scores, with gains of 17.2\% for PR and 20.9\% for PRV2$_{E}$. Remarkably, the boundary F1 scores even surpass those achieved during zero-shot inference, demonstrating the model's strong ability to predict accurate depth at object boundaries. Additionally, we evaluate different methods for generating pseudo labels. Although the offline approach shows similar performance, it eliminates the need for forward passes through the teacher model during training.

\begin{table}[t!]
    \centering
    \scalebox{0.65}{
    \begin{tabular}{C{0.3cm}|*{4}{C{0.85cm}}|*{3}{C{1.2cm}}}
        \toprule
        &\multicolumn{4}{r|}{Method}  & \textbf{RMSE} & \textbf{\#param.} & \textbf{T}(s) \\
        \midrule
        &\multicolumn{4}{r|}{Coarse Baseline} & 1.289 & - & - \\
        \midrule
        & F2C & E2E & C2F & NP \\
        \midrule
        \ding{172} & \checkmark & & & & 1.201 & 27.5M & 0.08s \\
        \ding{173} & \checkmark & & & & 1.214 & 70.2M & 0.38s \\
        \ding{174} & \checkmark & \checkmark & & & 1.184 & 27.5M & 0.08s \\
        \ding{175} & \checkmark & \checkmark & \checkmark & & 1.041 & 47.0M & 0.32s \\
        \ding{72} & \checkmark & \checkmark & \checkmark & \checkmark & 1.003 & 47.0M & 0.32s \\
        \midrule
        \ding{177} & \multicolumn{4}{r|}{w/o GDU} & 1.137 & 34.5M & 0.19s \\
        \ding{178} & \multicolumn{4}{r|}{replace GDU with PatchRefiner fusion} & 1.202 & 47.0M & 0.32s\\
        \midrule
        \ding{179} & \multicolumn{4}{r|}{AP, only load encoder} & 1.029 & 47.0M & 0.32s\\
        \ding{180} & \multicolumn{4}{r|}{w/o ImageNet pretraining} & 1.059 & 47.0M & 0.32s\\
        \bottomrule
    \end{tabular} 
    }
    \caption{\textbf{Ablation study of the framework on UnrealStereo4K.} F2C and C2F denote the fine-to-coarse and coarse-to-fine module in the bi-directional fusion module, respectively. E2E and NP are short for end-to-end training and noisy pretraining. Time: average inference time of the refiner branch for one image.}
    \label{tab:arch_ablation}
\end{table}

\subsection{Ablation Studies and Discussion}

We ablate and discuss the contributions of individual components proposed for PRV2. We employ the MobileNet in the refiner branch and adopt $P=16$ patches for clarity and ease of comparison. The inference time benchmarks are performed on a single NVIDIA A100 GPU.

\noindent \textbf{Framework Design:} As shown in Tab.~\ref{tab:arch_ablation}, we start with a baseline framework (\ding{172}) in which we only substitute the base depth model in the refiner branch in PR with the lightweight encoder. While the inference time and model size are drastically reduced, quality is also degraded by a large margin. Simply adding more parameters to scale up the F2C cannot improve the performance, as shown in \ding{173}. Adopting an end-to-end training strategy can help improve the model performance (\ding{174}). With the help of our proposed C2F that denoises the refiner features effectively, as shown in Fig.~\ref{fig:c2f_motivation}, the model's RMSE is reduced by 12.2\% while only introducing a satisfactory overhead (\ding{174}, \ding{175}). We also adopt different variants for C2F to evaluate the effectiveness. Firstly, we remove the GDU so that the C2F degrades to a simple bottom-to-top aggregation module (\ding{177}). While it can still improve the model performance, there is a large margin compared with the complete C2F module. Then, we replace the GDU with the fusion module used in F2C (\ding{178}). This results in a significant drop in performance. We argue this is due to the coarse features dominating the fusion process. The high-frequency information cannot be preserved correctly, leading to a performance on par with \ding{172}.

\noindent \textbf{Noisy Pretraining:} When equipped with the NP (\ding{72}), our model achieves the best performance with 22.2\% improvement over the coarse baseline in terms of RMSE. To prove our claim in the method section, we conduct the experiment by only loading the encoder part parameters after the NP process (\ding{179}). The discrepancy in performance indicates that the pretraining of C2F and F2C modules is also crucial for the model performance, which is often ignored in the current depth estimation community. Then, we discard the ImageNet~\cite{deng2009imagenet} pretrained parameters for the lightweight encoder and train the entire refiner branch from scratch (\ding{180}). The result validates our assumption that merely pretraining the encoder is not enough for better performance due to the lack of depth-aligned representation.

\section{Conclusion}
\label{sec:conc}

We presented \textbf{PatchRefiner V2}, an enhanced and efficient framework for high-resolution monocular metric depth estimation. Building on the strengths of the original PatchRefiner, PRV2 introduces a lightweight refiner branch, dramatically improving inference speed and reducing model size. With the novel Coarse-to-Fine (C2F) module and Noisy Pretraining strategy, our framework successfully mitigates the challenges posed by noisy features and the lack of pre-training of the refiner branch. Furthermore, we introduced the Scale-and-Shift Invariant Gradient Matching (SSIGM) loss to enhance boundary accuracy and improve synthetic-to-real transfer. Our framework significantly outperforms previous methods on the UnrealStereo4K dataset, achieving up to 9.2x fewer parameters and 10.7x faster inference. PRV2 also demonstrates considerable improvements in depth boundary delineation on real-world datasets like CityScape, ScanNet++, and KITTI, showcasing its adaptability and effectiveness across different domains and base models.

\appendix

\section{Dataset}
\label{sec:data}

\noindent \textbf{UnrealStereo4K (Synthetic, 4K):} The UnrealStereo4K dataset~\cite{tosi2021smd} consists of synthetic stereo images with a resolution of 2160$\times$3840 pixels, each paired with precise, boundary-complete pixel-wise ground truth. Images with labeling inaccuracies are excluded based on the Structural Similarity Index (SSIM)~\cite{wang2004ssim}, a process adapted from~\cite{li2023patchfusion,li2024patchrefiner}. Ground truth depth maps are computed from the provided disparity maps using specific camera parameters. Consistent with the splits suggested in~\cite{tosi2021smd,li2023patchfusion,li2024patchrefiner}, the experiments utilize a patch size of 540$\times$960 pixels for fair comparison.

\noindent \textbf{CityScapes (Real, Stereo):} The CityScapes dataset~\cite{cordts2016cityscapes} provides a diverse collection of urban scene images, segmentation masks, and disparity maps at a resolution of 1024$\times$2048 pixels. This dataset surpasses many in its domain in terms of image density, volume, and resolution~\cite{silberman2012nyu,song2015sunrgbd,schops2017eth3d,scharstein2014mid}. For our experiments, we use a standard patch size of 256$\times$512 pixels, primarily focusing on this dataset for testing our models following \cite{li2024patchrefiner}.

\noindent \textbf{ScanNet++ (Real, LiDAR, Reconstruction):} ScanNet++~\cite{yeshwanth2023scannet++} is an expansive dataset featuring high-resolution indoor images (1440$\times$1920 pixels), low-resolution depth maps from iPhone LiDAR sensors (192$\times$256 pixels), and high-resolution depth maps derived from laser scan reconstructions (1440$\times$1920 pixels). The low-resolution depth maps are employed during the training phase following~\cite{li2024patchrefiner}, with a selected patch size of 720$\times$960 pixels to accommodate the high-resolution setup.

\noindent \textbf{KITTI (Real, LiDAR):} The KITTI dataset~\cite{geiger2012kitti} includes stereo images and corresponding 3D laser scans of outdoor scenes, captured using equipment mounted on a moving vehicle. The RGB images are captured at a resolution of 376$\times$1241 pixels, paired with sparse ground truth depth maps. Adhering to the widely-used Eigen split~\cite{eigen2014mde}, we employ approximately 26,000 images from the left camera for training and 697 frames for testing. In line with the methodology of Garg et al.~\cite{garg2016unsupervised}, the images and depth maps are cropped to a resolution of 352$\times$1216 pixels for evaluation purposes, with a patch size of 176$\times$304 pixels used across the KITTI experiments.

\section{Ablation Study}
\label{sec:ablation}

\begin{table}[t!]
    \centering
    \scalebox{0.81}{
    \begin{tabular}{L{3cm}|*{2}{C{1.6cm}}}
        \toprule
        Method & \textbf{EdgeAcc}$\downarrow$ & \textbf{EdgeComp}$\downarrow$ \\
        \midrule
        SSI Loss~\cite{Ranftl2022midas,yang2024depthanything,li2024patchrefiner}  & 3.28 &  17.04 \\ 
        GM Loss~\cite{li2018megadepth} & \textbf{3.00} & 15.48 \\
        GMSSI Loss & 3.25 & \underline{14.89} \\
        \textbf{SSIGM} Loss (Ours) & \underline{3.03} & \textbf{14.70} \\
        \bottomrule
    \end{tabular}
    }
    \caption{\textbf{Ablation study of the loss design on CityScapes.} GMSSI indicates that we first calculate the gradient map of the prediction and pseudo label, align them via LSE, and then calculate the scale-and-shift invariant loss.}
    \label{tab:loss}

\end{table}

\noindent \textbf{Synthetic-to-Real Transfer:} We ablate the SSIGM loss on the CityScapes dataset and utilize the offline manner to generate pseudo labels. To achieve fair comparison with PatchRefiner (PRV1)~\cite{li2024patchrefiner}, we apply various types of losses on the vanilla PRV1 framework instead of the advanced PRV2 in this ablation study. As shown in Tab.~\ref{tab:loss}, directly adopting the gradient matching loss can significantly boosts the boundary metric. Combining the SSI and GM losses can combine the best of both worlds. Our experiments also indicate that the order of combination matters for a better trade-off. First adopting the scale-and-shift alignment and then applying the gradient matching works better compared with the reversed one (GMSSI).

\begin{table}[t!]
    \centering
    \scalebox{0.7}{
    \begin{tabular}{C{0.3cm}|*{4}{C{0.85cm}}|*{3}{C{1.2cm}}}
        \toprule
        &\multicolumn{4}{r|}{Method}  & \textbf{RMSE} & \textbf{\#param.} & \textbf{T}(s) \\
        \midrule
        &\multicolumn{4}{r|}{Coarse Baseline} & 1.289 & - & - \\
        \midrule
        & F2C & E2E & C2F & NP \\
        \midrule
        \ding{172} & \checkmark & & & & 1.118 & \textbf{51.7M} & \textbf{0.29s} \\
        \ding{173} & \checkmark & & & & 1.185 & 95.6M & 0.47s \\
        \ding{174} & \checkmark & \checkmark & & & 1.100 & \textbf{51.7M} & \textbf{0.29s} \\
        \ding{175} & \checkmark & \checkmark & \checkmark & & 0.985 & 72.1M & 0.57s \\
        \ding{72} & \checkmark & \checkmark & \checkmark & \checkmark & \textbf{0.947} & 72.1M & 0.57s \\
        \bottomrule
    \end{tabular} 
    }
    \caption{\textbf{Ablation study of PRV2$_E$ on UnrealStereo4K.} F2C and C2F denote the fine-to-coarse and coarse-to-fine module in the bi-directional fusion module, respectively. E2E and NP are short for end-to-end training and noisy pretraining. Time: average inference time of the refiner branch for one image.}
    \label{tab:arch_ablation_prv2e}
\end{table}

\begin{table}[t!]
    \centering
    \scalebox{0.7}{
    \begin{tabular}{C{0.3cm}|*{4}{C{0.85cm}}|*{3}{C{1.2cm}}}
        \toprule
        &\multicolumn{4}{r|}{Method}  & \textbf{RMSE} & \textbf{\#param.} & \textbf{T}(s) \\
        \midrule
        &\multicolumn{4}{r|}{Coarse Baseline} & 1.289 & - & - \\
        \midrule
        & F2C & E2E & C2F & NP \\
        \midrule
        \ding{172} & \checkmark & & & & 1.095 & \textbf{226.9M} & \textbf{0.38s} \\
        \ding{173} & \checkmark & & & & 1.151 & 270.9M & 0.61s \\
        \ding{174} & \checkmark & \checkmark & & & 1.089 & \textbf{226.9M} & \textbf{0.38s} \\
        \ding{175} & \checkmark & \checkmark & \checkmark & & 0.946 & 245.8M & 0.62s \\
        \ding{72} & \checkmark & \checkmark & \checkmark & \checkmark & \textbf{0.883} & 245.8M & 0.62s \\
        \bottomrule
    \end{tabular} 
    }
    \caption{\textbf{Ablation study of PRV2$_C$ on UnrealStereo4K.}}
    \label{tab:arch_ablation_prv2c}
\end{table}

\noindent \textbf{Framework Design:} We present more ablation studies about framework design based on PRV2$_E$ and PRV2$_C$ as shown in Tab.~\ref{tab:arch_ablation_prv2e} and Tab.~\ref{tab:arch_ablation_prv2c}, respectively. we start with a baseline framework (\ding{172}) in which we only substitute the base depth model in the refiner branch in PR with the lightweight encoder. While the inference time and model size are drastically reduced, quality is also degraded. Simply adding more parameters to scale up the F2C cannot improve the performance, as shown in \ding{173}. Adopting an end-to-end training strategy can help improve the model performance (\ding{174}). With the help of our proposed C2F that denoises the refiner features effectively, the model's RMSE is reduced while only introducing a satisfactory overhead (\ding{174}, \ding{175}). When equipped with the NP (\ding{72}), our model achieves the best performance.

{
    \small
    \bibliographystyle{ieeenat_fullname}
    \bibliography{main}
}


\end{document}